# Real-Time Predictive Maintenance using Autoencoder Reconstruction and Anomaly Detection

Sean Givnan[1], Carl Chalmers[1], Paul Fergus[1], Sandra Ortega[1] and Tom Whalley[2]
[1]School of Computing and Mathematical Sciences, Liverpool John Moores University, Byrom Street, Liverpool, L3 3AF, UK.
[2]Central Group, Kitling Road, Knowsley Business Park, Liverpool, Merseyside, L34 9JA.

*Abstract* - **Rotary machine breakdown detection systems are outdated and dependent upon routine testing to discover faults. This is costly and often reactive in nature. Real-time monitoring offers a solution for detecting faults without the need for manual observation. However, manual interpretation for threshold anomaly detection is often subjective and varies between industrial experts. This approach is ridged and prone to a large number of false positives. To address this issue, we propose a Machine Learning (ML) approach to model normal working operation and detect anomalies. The approach extracts key features from signals representing known normal operation to model machine behaviour and automatically identify anomalies. The ML learns generalisations and generates thresholds based on fault severity. This provides engineers with a traffic light system were green is normal behaviour, amber is worrying and red signifies a machine fault. This scale allows engineers to undertake early intervention measures at the appropriate time. The approach is evaluated on windowed real machine sensor data to observe normal and abnormal behaviour. The results demonstrate that it is possible to detect anomalies within the amber range and raise alarms before machine failure.**

## 1. Introduction

Rotary machine breakdowns are a significant problem in industry [1]. Breakdowns increase costs to the organisation in terms of compensation, reputation and fines incurred from missing production targets [2], [3]. Electric rotary machines account for over 40% of all global electricity consumption [4], which demonstrates their high dependence worldwide. When rotary machines fault and cause prolonged unplanned shutdowns, this often has a detrimental effect on business continuity. The complete supply chain is affected and the source organisation can often incur significant penalties.

There are three primary types of machine maintenance, reactive, preventative, and predictive. Each strategy has different benefits, costs, and limitations. An increasing number of organisations are moving towards predictive maintenance in order to reduce costs and maintain efficiency. This approach is underpinned by a variety of different technologies which include Internet of Things (IoT), Deep Learning (DL) and communication technology. The combination of IoT for remote data sensing, and the advancements in sensor analytics are core components of the Industry 4.0 revolution. By utilising the self-monitoring aspects of Industry 4.0, machinery can be monitored in real-time for the assessment of ongoing health and the detection of imminent faults [5]. These strategies are important as mechanical breakdowns are inevitable across all industries. The aim is to mitigate the risk of avoidable catastrophic breakdowns to production and manufacturing facilities [3].

In reactive maintenance, machinery failure causes a reactive response to resolve the fault. This method is not suitable for many industrial practices as unexpected downtimes can be costly [6]. Cost cutting exercises, such as the removal of regular maintenance, are often the cause of major faults [7], [8]. Conversely, preventative maintenance routinely tests and maintains machinery to detect faults and prolong the health of equipment [9], [5]. While, this helps to mitigate the negative impact caused by reactive maintenance, this strategy can also be costly in terms of time spent to maintain machinery, unnecessary hardware replacement and factory downtime.

Automating predictive monitoring using IoT and ML is an approach currently being considered which will facilitate controlled shutdowns around production as and when they are required. This approach prolongs machine health and reduces the occurrence of unplanned downtime [10]. By combining preventive maintenance with IoT and ML this further ensures that machine parts are only replaced when required, which is not always the case in many traditional preventative maintenance strategies [11], [12]. The difficulty with ML approaches is the lack of data containing a balanced representation of both normal and abnormal behaviour and the specific classes of fault that exist. More importantly the performance of many approaches are evaluated under lab conditions.

This paper proposes a state-of-the-art solution to monitor normal rotary machine behaviour obtained from infield motors operating under normal working conditions and highlight the occurrence of anomalous behaviour. Taking into account that anomalous data is a minority, as faults only occur sporadically, the approach taken in this paper focuses on anomaly detection rather than classification to predict when faults might occur. The approach utilises Stacked Autoencoders (SAEs) to extract features from raw signals found in normal operational behaviour and uses these to identify anomalies (i.e., feature values that deviate from what is considered normal behaviour). The paper primarily focuses on the detection of anomalies in real-time using an edge device which reduces the overall transmission of data sent for analysis. The edge device detects anomalies and decides what data to transmit and discard. The analysis in this paper is concentrated on electric rotary machines, (Motors, Pumps and Fans), as these have a significant impact on production, when failure occurs [13].

The remainder of this paper is organised as follows. The background is presented in section 2 and provides a discussion on current approaches and their associated limitations. The methodology is presented in section 3 which includes a discussion on data collection, data pre-processing and the configuration of the SAE along with the evaluation metrics used in the experiment to calculate the reconstruction error for anomaly detection. The results obtained from the model are presented in section 4 and discussed in 5 before the paper is concluded and future work presented in the section 6.

## 2. BACKGROUND AND RELATED WORKS

Various methods and approaches have been proposed to predict upcoming faults in rotary machines and facilitate predictive maintenance. Manufactures require a system that can distinguish between fault data and known healthy data in real-time [14]–[16]. For example, [17] developed an Automatic Weather Station system (AWS) to collect sensor data and model normal behaviour to reduce downtime in sensor weather stations. The approach model's behaviour to identify outliers in the sensor data produced in the AWS system using anomaly detection. The abnormal readings where modelled using correlations in data trends and Successive Pairwise REcord Differences (SPREDs). A similar approach was used in [20], whereby sensors are used to monitor devices in order to measure power consumption, performance and utilisation metrics, to identify abnormalities.

In the case of predictive maintenance, the data is often asymmetric as discussed in [22]. The main concern highlighted in the paper is the bandwidth needed to transmit real-time sensor signals. To address this the paper proposes the Anomaly Detection based Power Saving scheme (ADEPOS). Modelling healthy data and detecting changes overtime is performed at the edge while only anomalies are returned thus saving bandwidth. Extreme learning machine boundary (ELM-B) is used to detect anomalies and utilises traditional Autoencoders (AEs) configured with start-up weights and biases randomly selected from a continuous probability distribution.

Addressing the asymmetric problem and the large amount of data generated by sensors [18] performs all compute on the edge without returning data. It proposes a fault detection and monitoring system using a long short-term memory (LSTM) to model faults generated by machines in real-time. The approach analyses and detects anomalies at the edge and does not transmit any data back to centralised servers. It is not clear from the paper how the inherent limitation of inferencing complex models on edge devices scales to more complex real-world problems. The paper reports perfect results for Precision and Recall using simulated tests but there is no information on how the model performs when deployed in real-world environments. There is also a fundamental flaw in that the simulated data does not account for the inherent imbalance in machine generated data under normal working conditions.

AE are more recently being applied to model sensor data. An AE is a feed-forward neural network consisting of an input layer, at least one hidden layer, and finally an output layer. They are typically trained to reconstruct their own inputs, and therefore have the same number of inputs as outputs. A representation of an AE is displayed in figure 1, in which the model is used to check sensor data readings and determine an anomaly. Figure 1 displays a representation of an AE with three layers of hidden neurons. An AE with a lesser number of hidden neurons than the input/output layer is forced to learn complex features within the data, as the data is compressed into a lower dimensionality representation of itself, similarly, to learning from Principal Component Analysis (PCA) but from a nonlinear representation. These learnt features are used to reconstruct the input vector and can be used for different purposes, such as dimensionality reduction and image denoising.

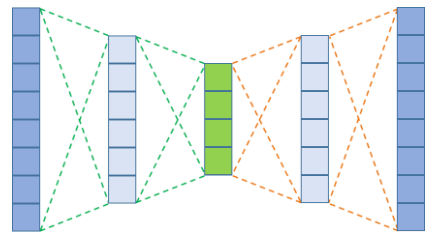

Figure 1: Autoencoder with two encoding layers displayed with green edges.

SAEs are combinations of multiple pretrained AEs that are individually trained [19]. By training an initial AE with a single layer, the model takes an initial input, such as a data window, and will reduce or 'encode' the input into a much smaller dimensionality, and 'decode' this layer to recreate the initial input, as the models output. This method allows relationships within the data to be found, by reducing the dimensionality allowing the model to create its own features prevalent within the data.

AEs work well on high-level features but have limited effect on signals which exhibit complex features. SAEs on the other hand allow more complex features to be extracted. Features present within the data can be found without the need for expert knowledge in signal processing. In [21] SAEs are implemented to determine the health of bearings using the NASA bearing dataset [22]. The results show that the method could achieve 100% detection for bearing faults.

Industry 4.0 and the idea of connecting sensors is a relatively new initiative. Combining this with ML is even newer. Nonetheless the background highlights a need for more complex analytics to optimise workflows and maintain the uptime of machinery used in manufacturing processes. This paper builds on the body of work in this area and presents the results of an industrial trial and assess the applicability of ML models and their ability to detect anomalies.

## 3. METHODOLOGY

A study was conducted in partnership with Central Group PLC in Liverpool UK. The dataset contains, time series sensor data, which was provided by a large steel manufacturer. The data was obtained from a critical machine used in their production line and contains both normal and fault data.

### A. Data Collection and Description

The sensor data was obtained from an electrical rotary machine that adds tension to feed product through production.

The machine is a critical component, and therefore has been pre-installed with sensors to determine the speed, current and voltage of the motor in real-time.

The data provided includes a large proportion of known healthy data. The manufacturing company also provided fault data, with their engineer reports diagnosing the reason for the failure. The data collected from rotary machines differ depending on the direct application of the equipment. Some rotary machines will have access to a drive, called a drive system, this will allow measurements to be taken such as: speed, frequency, current, and voltage. A drive is an electronic device that regulates the electrical energy delivered to a motor. The drive provides the motor with various amounts of electricity and regulates the frequency to control the motors speed.

Measurements taken from the motor include current, status bits, 1st up fault, 2nd 16 faults, armature current ref, actual armature current, speed reference, actual motor speed, field current ref, actual field current and actual armature voltage. Each of the 11 data channels is sampled at 100Hz and sample readings can be seen in figure 2.

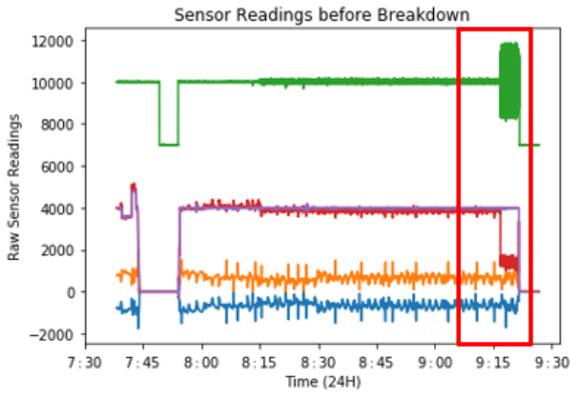

Figure 2: Time Series Sensor Dataset displaying at 1-hour 45-minute window of data, this is directly before the Catastrophic Failure highlighted in red.

For the model, a single measure of sensor data was used, Actual Field Current, as this was most consistent across the whole of the normal data and was the single most the prominent feature indicating the upcoming fault, as seen in figure 2. Note that Field Current Ref also shows significant change during the fault indicating that this channel strongly correlated with Active Field Current as shown in figure 3. As such we decided to use Actual Field Current only.

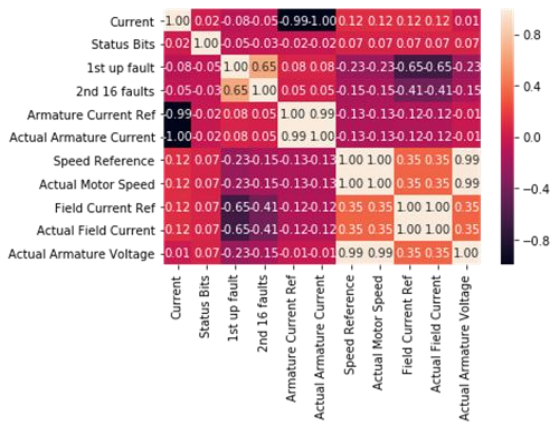

Figure 3: Correlation Matrix to test relationships between incoming sensor signals.

## B. Data Pre-processing

The data pre-processing begins by dividing the data into two sections, which includes pre and post failure. The primary focus is the detection of catastrophic failure. Figure 2 shows two hours of sensor monitoring. The data consists of 1 hour 45 minutes of normal operation (no fault) followed by 15 minutes of pre-fault data as highlighted in the red box. During the last 15 minutes it clear to see distinct changes in Actual Field Current prior to the actual breakdown which occurred at ~9:20 am. All other signals show limited variation as such they are discarded.

Actual Field Current has been identified as the channel of data that can best capture both normal and abnormal operation. For the purpose of this study Actual Field Current data from normal operation is extracted for machine learning modelling. A separate set of anomalous data (data leading up to the breakdown) is extracted from the signal data for testing the model.

The timeseries signals are segmented into various window sizes and used to model normal behaviour and detect anomalies. The number of hidden node parameters exponentially increases based upon the initial window size, therefore no more than 1000 samples (10 second window) will be used for analysis. In this study manual feature extraction is not implemented. Each of the windows are inputted through an ML model and features are extracted automatically as discussed in the next section.

## C. Machine Learning and Modelling

The framework aims to identify abnormal behaviour in rotary motors using the Actual Field Current signal. The signal is fed into an anomaly detection model and abnormal behaviours identified by calculating the reconstruction error. Anomaly detection can use the reconstruction capabilities of a pre-trained ML model, to identify when sensor readings become abnormal and therefore could contain valuable information.

The overall process applied in this paper seeks to denoise the sensor data based on known healthy data, and in the process determine anomalies from abnormal readings when new untrained data is tested using an AE. The abnormal readings will give further insights to possibly identify when a sensor reading starts to become abnormal based on known normal data.

The architecture of an AE consists of an encoder and a decoder. The encoder reduces the dimensionality of the data by learning patterns and features of the data. The connections between the input and hidden nodes are the encoding layer of the AE. The decoding layer is the connections between the hidden layer and the output nodes, this segment of the AE uses the features learnt to recreate the input from a lower-dimensional representation. An AE consists of neurons, these neurons take inputs and apply a weight and bias to each input, then aggregated and applied to an activation function, such as a sigmoid function.

$$h_{W,b} = f\left(\sum_{i=1}^{m} W_i x_i + b\right)$$

Equation 1: Neuron computational formula for updating each weight to approximate the input. W- weight, b – bias, i – number of input neuron, m – max input neuron, x – input vector, f – activation function.

Subsequent neurons learn smaller (dimensional) representations of the inputs by adjusting the weights to reconstruct each input with a closer approximation as these weights are updated. The result allows an N-dimensional dataset to be reconstructed using P neurons as a representation of the dataset (N > P). The AE allows an edge detection model to be inferenced in real-time to determine any abnormal behaviours from the rotary machine.

Backpropagation is used to minimise the overall cost function throughout the network, here this is the overall reconstruction error using mean squared error. This is achieved by optimising the network and calculating the gradient of the cost function with respect to the weights and biases, then adjusting the weights and biases at each layer to improve the overall reconstruction error. Backpropagation trains the weights within the network by optimising each weight based upon the training sample at time t, with weight $\omega_{ij}$, and is updated using equation 2.

$$\Delta \omega_{ij}^t = -\epsilon \frac{\partial E}{\partial \omega_{ij}} + \alpha \Delta \omega_{i,j}^{t-1}$$

Equation 2: Backpropagation for the model network, to update each individual weight. This updates each weight within the network to decrease the overall reconstruction error. w - Weight, t - time, $\epsilon$ - learning rate, E - Error function.

The network is individually trained layer by layer. This means that backpropagation was not conducted on the final model, as each layer is trained individually. Backpropagating across the whole network would produce a Denoising AE (DAE) rather than a SAE, and this is an important distinction to make between the two methods. Figure 4 outlines the method for producing individual models to create a SAE from individual AEs.

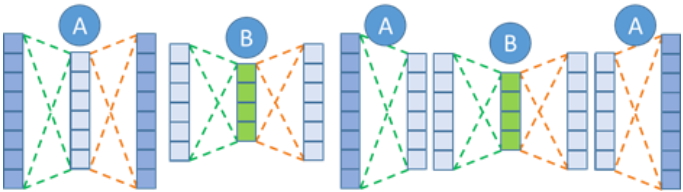

Figure 4: Stacked Autoencoder (SAE) architecture, consisting of two single models trained separately, which the hidden layer of model A being the input layer of model B. To the right, the model is shown in full (A-B-A).

Figure 4 displays an example of a SAE model, in which the models are individually trained, and the weights from each model are gathered and collated to create a final model. This architecture is different to the DAE, as this model trains using backpropagation throughout the network, therefore each subsequent layer is trained to minimise the overall error. SAEs use separate models to train a single hidden layer, these hidden layers learn features from the input data to recreate the output layer. The subsequent hidden layer is then used as input to train a further model. This results in the first model learning simple features of the training dataset, the second model learns more complex features based on the initial model's output from the features. This allows a more complex model to be created, as the model forces the data dimensionality to reduce, in turn determining more complex features in the data.

Figure 5 shows the Framework design to reduce the overall storage and transmission of real-time sensor data to only include data that signifies a potential issue with the machine. When a set of sensor data does not fit the pre-trained profile of the healthy dataset, the data is sent for further analysis to determine whether an upcoming fault is imminent. This addresses a significant issue with cloud storage. Transmission of large volumes of data using 4G/5G networks and storing raw data for further processing is inefficient, as a substantial proportion of sensor data will be classed as healthy. This healthy data does not contain valuable information, as the machine is known to be in full working order. Rather, it is better to store data indicative of an upcoming fault which will reduce the overall analysis needed to be conducted for real-time monitoring. This will reduce the computing power, storage and bandwidth needed in the initial analysis.

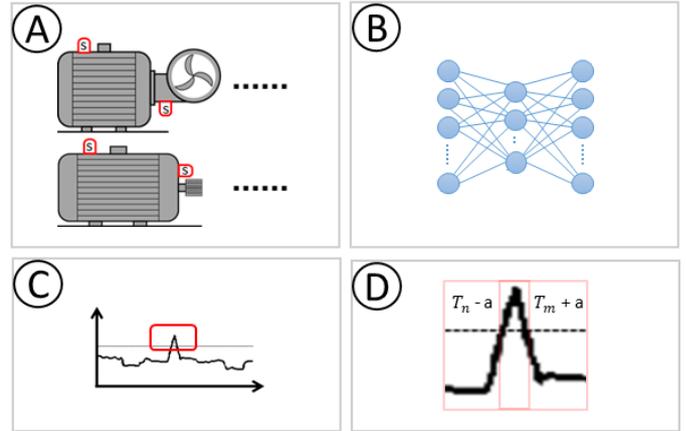

Figure 5: Overall Framework design for anomaly detection on the edge for data transmission.
A – Rotary Machines of various kinds with IoT Sensors attached delivering sensor data to an IoT device
B – IoT device is embedded with a pre-trained Stacked Autoencoder with a set of rules to investigate anomalies based on the reconstruction of sensor data.
C – Abnormality is detected by the IoT device.
D – IoT device transmits a window of T-minutes before the anomaly was detected including up to T-minutes after the final anomaly is detected by the threshold.

### D. Stacked Autoencoder Configuration

The model is trained using different windows size configurations. The first model uses a 500-observation window of sensor data equal to 5 seconds. In this model the hidden layer contains 300 neurons which ensures enough structure is kept from the initial data and sufficient features can be extracted. The output from the hidden layer connects to the output layer which contains 500 neurons. This model is trained to approximate the input on the output layer. Training continues until an optimal reconstruction error is achieved. When training in this model is complete the hidden layer (features) is used as input to a second AE which again is trained to approximate the input on the output layer. In the second model the hidden layer is reduce to 200 nodes. This configuration continues until the final model contains 70 hidden neurons which after training constitutes our final set of features (70 features).

Each individual model is built and combined to form an overall model, using the individual AEs to create an SAE. The SAE has an overall Layer structure as shown below.

$$Model\ 1: 500 - 300 - 500$$

$$Model\ 2: 300 - 200 - 300$$

$$Model\ 3: 200 - 120 - 200$$

$$Model\ 4:\quad 120 - 70 - 120$$

The final set of features (70) obtained from the SAE is used to reconstruct the output as shown in model 5.

$$Model\ 5:\quad 70 - 50 - 500$$

Using a custom loss function, the model is able to compare the initial window of data (size 500) with the output layer generated in model 5. The loss function for models 1-4 determine the difference between the model input and output, and minimises the loss. This is not be suitable for model 5, so the model takes both the initial training data, and the subsequent features of the model at the 70 layer as shown below:

$$500 - 300 - 200 - 120 - 70$$

The model evaluates the loss by comparing the models input and output based on the reconstruction of the data using the 70 features. This is achieved using a squared distance measure as seen in equation 3. This equation calculates the vector distances to ensure a close reconstruction has been met. Historical healthy data is used to train the model above to recognise features present during normal operation. Once trained the model will be unable to recognise unusual data points, as such the reconstruction error will higher. This is used to detect anomalies in machine operation.

$$SD = Mean((A - B)^2)$$

Equation 3: Squared Distance measure between the input and output vectors, used to train the final model in the Autoencoder. With A being the initial input window, and B being the total model output.

The Squared Distance measure in Equation 3 is used to train model 5 and compare the original input using the expanded output generated by model 5. This is a custom loss function added to the model, as it does not take the models input features as a measure for the loss function, it uses the initial data window to evaluate the loss function, and the corresponding features of these same window as the input of the model.

### E. Evaluation Metrics

The models were trained using Tanh for both encoder and decoder. The loss value was computed using mean squared error and the optimiser of the model was Stochastic Gradient Descent (SGD). Gradient Descent (GD) is an iterative process to calculate the optimal value of a function by minimising the cost parameter. SGD uses the same process but randomly selects a small proportion of samples from the training dataset and calculates the gradient. Each parameter is updated from SGD by computing a small number of training samples at each iteration. This allows the model to train using a large dataset which computationally expensive using GD. The parameters are updated using the following equation:

$$\theta = \theta - \alpha \nabla_\theta J(\theta; x^{(i)}, y^{(i)})$$

Equation 4: SGD formula for updating parameters based upon a small sample of data taken at each iteration. $\alpha$ - learning rate of the SGD, x - input vector, y - output vector.

The mean squared error (MSE) loss function calculates the average of squared differences between the given input values, and the predicted output values, and minimises the error across the whole training dataset. This ensures a minimum error across the training data set. MSE is calculated by minimising the following formula shown in Equation 5.

$$MSE = \frac{1}{n}\sum_{i=1}^{n}(Y_i - \hat{Y}_i)^2$$

Equation 5: MSE formula for calculating a minimum error between a given input and output vector. n – Total number of samples, $Y_i$ - input vector, $\hat{Y}_i$ - target output vector.

The reconstruction error is vital in determining unusual activity from the sensor readings. Significant differences will highlight data not contained in the healthy dataset. Therefore, this system will act as a filter, to remove any unneeded data, when the system is functioning under normal operation, allowing any data outside of this norm to be identified as an anomaly. This data can then be tested to determine a possible diagnosis, by allowing transmission of data to an outside source for example an engineer.

The initial parameters for 'unusual' readings will be based upon the summary statistics of the healthy dataset. This means determining a percentile region, from which the reconstruction of the healthy dataset contains approximately 99.99% of the healthy data. Therefore, this region will cover almost every healthy data point, except for a handful of windowed sensor readings containing much higher reconstruction error. Further boundaries can be set, using a traffic light system, with green, amber and red showing healthy, unhealthy, and critical regions successively. This system is designed to be flexible to accommodate stricter analysis if the targeted rotary machine requires special care. The thresholds can be altered to stricter settings, depending on the needs of the individual, which will allow more data to be abnormal, subsequently transmitting more data for classification of the fault.

Finally, an observation made during training and testing found that when sensor data is taken at a resting point, a set of hidden nodes within the model converge to predict this resting period. However, this reduces the overall predictive power of the model for non-resting data. To combat this, no resting points are filtered as the model is unable to predict these data points. This ensures that the model has better convergence during normal running periods.

### 4. RESULTS

The experiments in this section use blocks of data (windows) between 250 – 1000 from the Actual Field Current channel during normal operational conditions. Each model in the SAE is independently trained, based upon the previous models condensed hidden layer. Hidden layers are fed as inputs into subsequent models.

Data reconstruction for each window is used to determine thresholding scores. This provides the basis of the anomaly detection, as any unusual patterns in the data, should be considered anomalies and require further analysis.

As the healthy data is used to train the model, the reconstruction of this data will be precise. Therefore, using this particular model and reapplying this to the healthy training data allows a threshold to be determined based on the reconstruction error calculated. This will be the basis of the anomaly detection, as the threshold will allow only the most extreme cases.

### A. Window - 500

Figure 6 shows the reconstruction error when the Actual Field Current is run through the SAE model. Using the reconstruction error, three thresholds are set, these provide a safety measure and ensure anomaly data is detected appropriately.

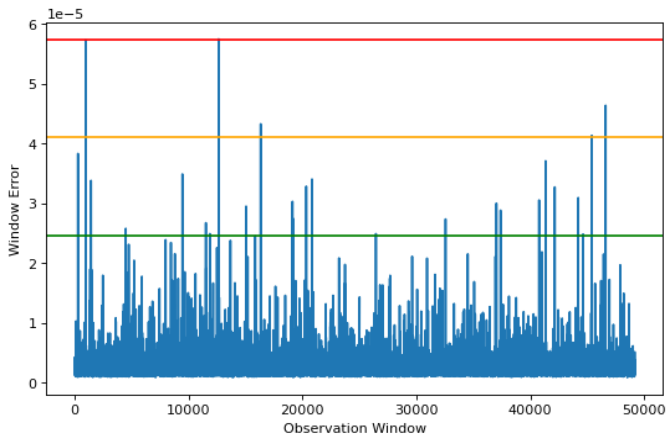

Figure 6: Reconstruction Plot using SAE model, on known healthy data, using MSE as the indicator. The Red line denotes the upper threshold based upon the maximum error from our model. The Green threshold is based upon the 99.95 percentile The Orange threshold is the region between the two thresholds.

The thresholds in figure 6 allow regions to be created, based upon the reconstruction capabilities from the SAE. Having three levels for the threshold provides more flexibility, as a small window of data may show a localised anomaly, compared to a period of windows showing anomalies leading to a failure. The red line in figure 6 shows the upper limit, which provides a suitable starting point for the anomaly detection. The green region also allows for safeguarding, so data on the verge of being an anomaly, is also detected.

Figure 7 denotes the reconstruction of unhealthy data, which in our dataset occurred on day 19. This data represents information leading up to the fault. The plot shows a 2-hour window before the fault occurred. In this instance the threshold is reached, and maintained above the red line. This occurred 70 minutes before the critical failure and provides engineers with an opportunity to perform predictive maintenance before catastrophic failure occurs.

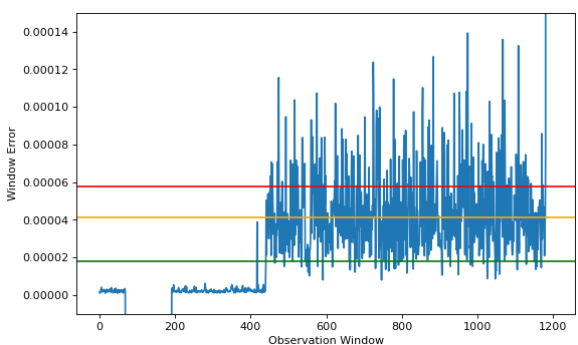

Figure7: Reconstruction Plot using SAE model, using a data window of 500, on known fault data, using MSE as the indicator. The lines denote the previous threshold based upon the healthy data used to train the model.

A window consistently above the green threshold would be a cause for concern, as the region below contains the error for 99.95% of all training data. Therefore, the model has identified a sudden change behaviour indicating an anomaly has occurred.

You will note in figure 7 that the signal between 65 and 190 are missing. This is because the production line has been stopped which is a frequent occurrence. In our approach the data is bypassed by setting the error to -0.001 which distinguishes the data from normal values. This ensures that a stoppage is not detected as an anomaly.

### B. Window – 250

The investigation will compare a further two window sizes, at 250 samples and 1000 samples. The same structure of model was used for both the 250 and 1000 sample window, which means a reduction of ~40% between layers to ensure sufficient information is retained, with the loss function evaluating the final layer.

Firstly, the 250-sample window was used to train the SAE. The SAEs were arranged in the same way as the 500-sample window but with the following configuration:

$$250 - 150 - 100 - 75$$

The main aim was to compare the known fault to determine whether increasing, or decreasing, the window size adversely affects the overall model performance, or possibly increases the detection. Figure 8 shows the final reconstruction window based on the performance of the SAE for unseen, abnormal data. The model using a data window of 250 produces similar result to the model trained with a window of 500. The vast number of observations fall within the green region, these observations would be considered normal. Reconstruction error spikes above the red threshold are prevalent, meaning that the machine would be showing signs of abnormal behaviour, however there is a standout difference in performance between the two models. Using a 500-window model has shown that the abnormal data is consistently above the green threshold, and regularly above the red threshold, which is desirable over the 250-window model.

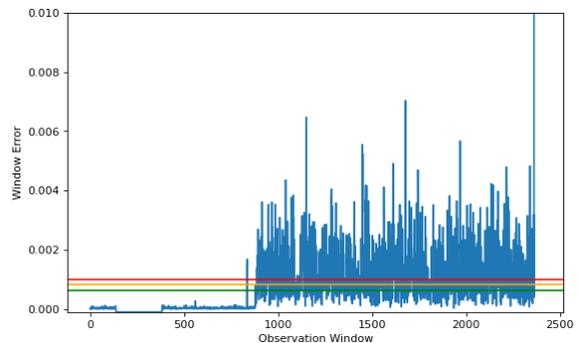

Figure 8: Reconstruction Plot using SAE model, using a data window of 250, on known fault data. The lines denote the threshold based upon the healthy data used to train the model, starting at the 99.95% percentile, for the green threshold.

As the reconstruction error in the 250-window model was better for the abnormal data, which is not the aim, the hypothesis would be that increasing the models initial window size will also increase its ability to determine abnormalities. This also comes with some limits, as the model should be regularly testing to ensure a fast detection of anomalies, by increasing the window size, the model will therefore use more samples to determine a difference, as a result it will predict less often. This will provide a trade-off between the number of predictions made, and the time taken to determine a fault.

### C. Window – 1000

The reconstruction plot for the 1000 window model is shown in Figure 9. The plot shows a significant increase in the error received when reconstructing unhealthy data. Compared to the previous window, the model has significantly differentiated between the abnormalities, by using 10 second windows of data within the SAE model.

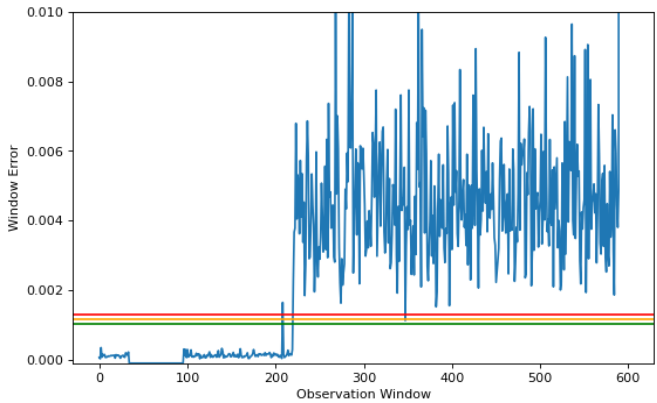

Figure 9: Reconstruction Plot using SAE model, using a data window of 1000, on known fault data. The lines denote the threshold based upon the healthy data used to train the model. The same level of thresholds were used in this model.

The results show a significant improvement over previous window sizes. The model is much stricter in determining anomalies, when compared to both the 250 and 500 data window models. The thresholds are much smaller, and closer together. This means the training data is better represented within this model, as the maximum threshold (red threshold) is far more stringent in this model than previous experiments.

## 5. DISCUSSION

This study presented a SAE anomaly detection model to identify abnormal behaviour in rotary machines. Based on the results, the best reconstruction capabilities were from the 1000-window model. The model using 500 samples performed well in determining some possible abnormalities, making use of all three thresholds. This was in contrast to the 1000-window model which more sensitive to very small anomalies in the Actual Field Current channel. The 1000-window model may be more suitable for motors with small deviations in pattern, but still requiring 24/7 uptime, as the model takes a brute force approach, in which smaller deviations are always significant.

The 500-window allows the model to determine differences in normal and abnormal data, whilst the severity of the abnormality can also be determined. This will allow the model to transmit small segments of data, that may be abnormal. This provides a safety barrier, that can determine upcoming faults from the different thresholds a model uses. Which model to use is a trade off and is dependent on the sensitivity required. In cases where small deviations from normal behaviour need to be detected the 1000 model would be preferable. In others where there is a need for lower sensitivity levels the 500 model would suffice.

Figure 6 displays the reconstruction error of training data used to train our model for anomaly detection. We created three thresholds, in which the severity of each observed window could be determined, with any window outside of our most extreme normal data observation requiring further investigation. Having 3 levels of thresholds allows the severity of the observed windows to be determined. The data below the green threshold will be regarded as normal behaviour, and therefore requires no further action. Data windows falling between green and amber thresholds will be seen as abnormal but not urgent. This category provides engineers with an opportunity for undertaking predictive maintenance. Data failing between amber and red is severely abnormal and requires immediate intervention.

The approach provides an important technique for defining thresholds that are personalised to particular rotary motors. This allows our model to be flexible, as the threshold is currently based on percentiles from our normal data rather than fixed point settings that often lead to a high number of false positives. More importantly these thresholds allow us to detect anomalies seventy minutes before breakdown. This provides critical information particularly in production lines which are fault sensitive and will likely have a significant impact on workflows that contain rotary machines.

This flexibility is vital to the overall framework for future developments, as a static value might be suitable, but having the ability to dynamically change this threshold based on the importance of a machine will allow more stringent classification tests to occur. The trade-off is between the amount of data sent for analysis, against the cost of a machine breaking down. An ideal scenario would be that all data is analysed, however filtering using the SAE, with a lower threshold will mean less data is analysed, but the cost of analysis through transmission and storage is less significant.

## 6. CONCLUSION AND FUTURE WORK

This paper proposed the use of a SAE to detect anomalies in signals received from rotary machine motors. The SAE provides an efficient data filter and requires no expert knowledge to determine when faults on rotary motors will likely occur. The SAE model provides a flexible system, for automatic threshold identification to identify normal and abnormal behaviour. The results are positive and the windowing strategy clearly shows a range of sensitivity levels are possible when evaluating deviations from normal operational behaviour. Once anomalies are detected fault signatures can be used by engineers to undertake maintenance. However, this does require expertise in signal processing and classification.

In our future work we will extend our anomaly detection algorithms and add new models trained on fault signatures to automatically classify the different types of faults that occur. The classification model will sit idle until data is provided from the anomaly detector presented in this paper. Future testing will be carried out using a simulation rig, consisting of a motor, gearbox and a pump setup, with access to vibration, temperature and drive sensors. The simulation rig will allow numerous faults be generated and signatures to be recorded and annotated.

There is an opportunity to include further sensors within the SAE and test the potential for transfer learning across different motor types and manufactures. Having a general-purpose model would be useful when used with a diverse range of motors.

It is also worth noting that the abnormal readings will assist in determining regions of sensor data which could be considered abnormal and therefore will define labels when creating further training data after a breakdown has occurred. The importance of this cannot be understated, as the sensor data is unlabelled, and access to a condition monitoring engineer with considerable experience in signal processing is not appropriate for every single signal diagnosis.

Overall, the model delivers an effective edge solution for detecting unusual behaviour, without the need for large data management, transmission and analysis. This model will identify erratic behaviour and indicate this to the operator of the rotary machine, with the opportunity to analyse what this anomaly signifies, and how this may affect production in the short term. This model provides a practical solution to determine localised anomalies on edge devices, whilst allowing information rich data to be found and used for further analysis.